\documentclass[conference]{IEEEtran}
\IEEEoverridecommandlockouts
\usepackage{cite}
\usepackage{amsmath,amssymb,amsfonts}
\usepackage{algorithmic}
\usepackage{graphicx}
\usepackage{textcomp}
\usepackage{caption}
\usepackage{subcaption}
\usepackage{stmaryrd}
\usepackage{url}

\usepackage{xcolor}
\def\BibTeX{{\rm B\kern-.05em{\sc i\kern-.025em b}\kern-.08em
    T\kern-.1667em\lower.7ex\hbox{E}\kern-.125emX}}
\begin{document}

\title{Semi-Supervised Graph Attention Networks for Event Representation Learning\thanks{This work was supported by National Council for Scientific and Technological Development (CNPq) [process number 426663/2018-7] and The São Paulo Research Foundation (FAPESP) [process number 2019/25010-5 and 2019/07665-4]}}

\author{\IEEEauthorblockN{Joao Pedro Rodrigues Mattos}
\IEEEauthorblockA{\textit{Institute of Mathematics and Computer Sciences (ICMC)} \\
\textit{University of S\~ao Paulo (USP)}\\
S\~ao Carlos, Brazil \\
joao\_pedro\_mattos@usp.br}
\and
\IEEEauthorblockN{Ricardo M. Marcacini}
\IEEEauthorblockA{\textit{Institute of Mathematics and Computer Sciences (ICMC)} \\
\textit{University of S\~ao Paulo (USP)}\\
S\~ao Carlos, Brazil \\
ricardo.marcacini@icmc.usp.br}
}

\maketitle

\begin{abstract}
Event analysis from news and social networks is very useful for a wide range of social studies and real-world applications. Recently, event graphs have been explored to model event datasets and their complex relationships, where events are vertices connected to other vertices representing locations, people's names, dates, and various other event metadata. Graph representation learning methods are promising for extracting latent features from event graphs to enable the use of different classification algorithms. However, existing methods fail to meet essential requirements for event graphs, such as (i) dealing with semi-supervised graph embedding to take advantage of some labeled events, (ii) automatically determining the importance of the relationships between event vertices and their metadata vertices, as well as (iii) dealing with the graph heterogeneity. This paper presents GNEE (GAT Neural Event Embeddings), a method that combines Graph Attention Networks and Graph Regularization. First, an event graph regularization is proposed to ensure that all graph vertices receive event features, thereby mitigating the graph heterogeneity drawback. Second, semi-supervised graph embedding with self-attention mechanism considers existing labeled events, as well as learns the importance of relationships in the event graph during the representation learning process. A statistical analysis of experimental results with five real-world event graphs and six graph embedding methods shows that our GNEE outperforms state-of-the-art semi-supervised graph embedding methods.
\end{abstract}
\IEEEpeerreviewmaketitle

\section{Introduction}

Event analysis from news and social networks is very useful for a wide range of social studies and real-world applications \cite{chen2020event}, such as the impact of epidemics, wars and urban violence, finance, elections, and sentiment analysis. Although machine learning methods have been recently explored to support event analysis, learning appropriate representations for event classification algorithms is a challenging task since events have different associated components, such as people, organizations, temporal and geographical information \cite{setty2018event2vec}. Traditionally, events have been represented using bag-of-words models, thereby focusing mainly on the textual information (e.g., terms and keywords) of the events \cite{allan2002topic}. However, such representation does not adequately capture more complex relationships between events and is often criticized for the lack of semantics \cite{aggarwal2018machine}. 

\begin{figure}[hptb]
    \centering
  \includegraphics[width=0.7\linewidth]{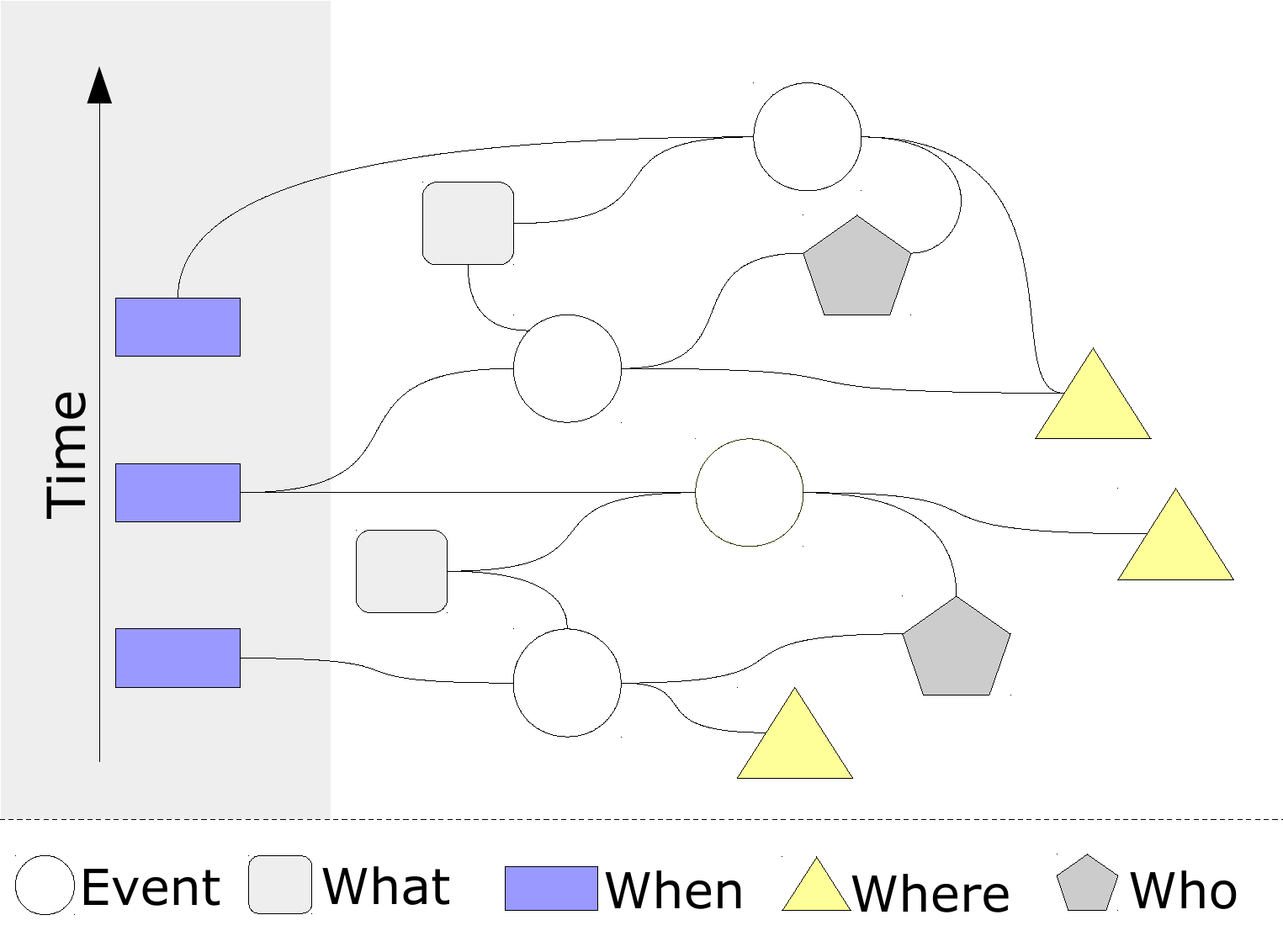}
  \caption{Example of an event graph \cite{dos2020two}.}
  \label{fig:event_network}
\end{figure}

Recent studies represent event components through journalistic ``w'' questions, such as what, when, where, and who \cite{hamborg2018giveme5w}. In this sense, events and their components can be explicitly represented using event graphs \cite{chen2020event} (Figure \ref{fig:event_network}), where edges indicate relationships between events and component vertices. Although this is a rich representation for event data, it poses new challenges for graph-based machine learning, which raises the following question: how to extract useful knowledge from event graphs and their complex relationships?

Graph embeddings methods have been used to learn latent features capable of capturing complex graph relationships, mapping each vertex in a low dimensional vector space \cite{cui2018survey, goyal2018graph, zhang2020deep}. This new representation is an embedding space used as input for several other tasks, such as event classification. For example, DeepWalk \cite{perozzi2014deepwalk} and Node2Vec \cite{grover2016node2vec} are methods based on short random walks to capture vertex neighborhood structures and learn features using a strategy similar to Word2Vec. Setty and Hose (2018) \cite{setty2018event2vec} proposed the Event2Vec method, which extends Node2Vec to respect event semantics through biased random walks. Graph Convolutional Networks (GCN) \cite{kipf2017semi} and Graph Attention Networks (GAT) \cite{velickovic2018graph} have also been extended for deep representation learning to extract high-level features from graphs. Despite recent advances, existing methods fail to meet the following important requirements for event analysis:

\begin{itemize}
    \item In event analysis tasks, a small amount of data can be labeled according to the user's feedback, such as the category or utility of the event. Previous graph embedding methods, such as DeepWalk and Node2Vec, are unsupervised and ineffective in integrating labeled data during the graph embedding process.
    \item Determining the importance level of the event components enables the extraction of complex patterns, for example, seasonal and geographical behaviors. Existing methods assume that such importance levels should be defined as parameters for edge weights between events and their components. However, it is impracticable for users to set these parameters manually.
    \item Although GCN and GAT are promising methods for semi-supervised graph embeddings, these methods are not suitable for graphs composed of different types of vertices and relationships, such as event graphs. Also, some vertices of event graphs have associated features, such as textual information. Both GAT and GCN are unfeasible to perform graph embeddings in these scenarios.
\end{itemize}

To address these limitations, we propose the GNEE (\textbf{G}AT \textbf{N}eural \textbf{E}vent \textbf{E}mbeddings), a new semi-supervised embedding method for event graphs using Graph Attention Networks (GAT) and Event Feature Regularization. Our GNEE innovates in incorporating both vertex labels and features to improve event representation learning. The key idea is to explore graph regularization to generate a textual-based representation for all vertex types, i.e., propagate semantic features extracted from event vertices. GNEE learns the final embeddings through graph neural networks with attention mechanisms. Our main contributions are two-fold:

\begin{itemize}
    \item We present a graph regularization framework to propagate existing textual features from event vertices to neighboring component vertices. We compute a semantic representation of the events through BERT-based neural language models. These models allow the generation of textual embeddings considering context information and pre-trained models from a large textual corpus. During the propagation of textual embeddings from event vertices to component vertices, a fine-tuning of the BERT-based representation is performed according to the structure of the event graph. The graph regularization step ensures that all vertices get a regularized feature, even non-text component vertices representing people's names, times, and locations. For example, the features of a location vertex will have semantic content similar to the event texts that occurred at that location.
    
    \item We propose a semi-supervised graph embedding process guided by both labeled vertices, regularized vertex features, and the graph topology. In particular, we explore a graph attention mechanism proposed by \cite{velickovic2018graph} to automatically learn different importance levels for each vertex, thereby automatically identifying when time, location, names of people, organizations, etc., are relevant for event embedding. Thus, GNEE performs graph embedding learning with an attention mechanism to obtain neural event embeddings according to the neighborhood structure of the component vertices. The expectation is that the model will identify which component event vertex are most important for event classification.
    
\end{itemize}

We carried out a thorough experimental evaluation on five real-world event datasets. Our GNEE was compared with two state-of-the-art semi-supervised graph embedding methods based on GAT and GCN and with three unsupervised graph embedding methods DeepWalk, Node2Vec, and Struc2Vec. A statistical analysis of the results reveals that the GNEE outperforms the previous GAT-based methods for neural event embeddings in classification tasks. Furthermore, GNEE proved to be competitive with existing methods based on GCN, DeepWalk, and Node2Vec.

\section{GAT Neural Event Embeddings}

Events are related to each other through a complex structure involving components such as people, organizations, locations, and particular time intervals. In this context, graphs allow identifying relationships between events and their components, which would not be possible using a representation model based only on texts, such as the traditional bag-of-words. In addition, it is also possible to enrich the representation by incorporating features and labels at each vertex, which is then used to improve the graph embedding process. However, even recent methods for semi-supervised graph embedding require that all graph vertices contain features, which is an unusual scenario in event graphs. Alternatively, such methods use the adjacency matrix itself as features, which discards essential event information. Ideally, a graph embedding method for events should (i) be semi-supervised to consider small sets of labeled events; (ii) consider existing features for event vertices, even if component vertices do not have associated features; and (iii) automatically learn the importance of the event components. Thus, we propose the GNEE method (GAT Neural Event Embeddings), which explores attention mechanisms and event features regularization to deal with these challenges.

Let $G = (V, E, W)$ be a graph where $V$ represents a set of vertices, $E$ a set of edges, and $W$ the weights between vertices and edges. We use a heterogeneous graph representation in which the vertices are composed of two types, $V = V_E \cup V_C$, where $V_E$ are event vertices and $V_C$ are component vertices. The latter represents information about people's names, organizations, locations, times, and other metadata related to the events. In our graph-based representation, the textual information for each event $v_i \in V_E$ is represented by a feature vector $\vec{g}_{v_i} \in \mathbb{R}^m$ in an $m$-dimensional space obtained by some text pre-processing technique, such as the BERT-based models (detailed later in this section). Moreover, the graph contains some labeled event vertices $V_L \subset V_E$ in $K$ classes $Y = \{1,. . . , K\}$, thereby forming a training set $\{(v_1, y_1), \dots, (v_n, y_n)\}$ for semi-supervised learning, with $ n = |V_L| $ and $y_j = k \in Y$.

The neural event embedding can be formulated as a mapping function $h : G(V,E,W) \rightarrow \mathbb{R}^d$ from vertices to a $d$-dimensional vector space (embedding space), where $d$ is a predefined parameter. Our proposed GNEE explores both the existing features and labels of the vertices to improve the embedding learning process, as well as the graph topology. GNEE first performs a feature regularization from the event vertices to component vertices, followed by a semi-supervised learning based on graph neural attention networks.

The textual information of the event dataset is used mainly to extract the components and relationships between events. After constructing the graph, most methods discard textual information and perform graph embedding using only the graph structure. Our GNEE incorporates textual information as a feature vector in the event vertices. We propose propagating such feature vectors to component vertices according to the network topology through a graph regularization framework. For example, if a component vertex representing a location is connected to multiple events, then that vertex must receive a feature vector similar to the event feature vectors.

We use the BERT neural language model \cite{devlin2019bert} to semantically represent the event textual data. Let $e = (t_1,...,t_k)$ be an event in which its textual information is a sequence of $k$ tokens. BERT explore a masked language modeling procedure, where one of the training objectives is the noisy reconstruction defined in Equation \ref{bertmlm},

\begin{equation} \label{bertmlm}
p(\bar{e}|\hat{e}) =  \sum_{j=1}^k m_j p(t_j,c_j)
\end{equation}

\noindent where $\hat{e}$ is a corrupted token sequence of the event $e$, $\bar{e}$ is the masked tokens, $m_j$ is equal to $1$ when $t_j$ is masked and $0$ otherwise. The $c_j$ represents context information for the token $t_j$, usually the neighboring tokens.

BERT uses a deep neural network based on the Transformers architecture to solve $p(t, c)$ of Equation \ref{bertmlm}. Typically, this strategy is reduced as conditional distribution modeling of the a token $t$ given a context $c$, according to Equation \ref{embeddings_bert},

\begin{equation} \label{embeddings_bert}
p(t | c) = \frac{exp(\mathbf{h}_c^{\top} \mathbf{w}_t)}{ \sum_{t'} exp( \mathbf{h}_c^{\top} \mathbf{w}_{t'} ) }
\end{equation}

\noindent where $\mathbf{h}_c$ is a context embedding and $\mathbf{w}_t$ is a word embedding of the token $t$. The term $\sum_{t'} exp( \mathbf{h}_c^{\top} \mathbf{w}_{t'} )$ is a normalization factor using all tokens $t'$ from a context $c$. Both embeddings $\mathbf{h}_c$ and $\mathbf{w}_t$ are obtained during the BERT pre-training stage from large textual corpus. In our approach, given an event $e = (t_1, ... , t_k)$, we compute the initial event semantic feature $\mathbf{g}_e$ by taking the average vector of all token embeddings $\mathbf{g}_e = \sum_{j=1}^k \frac{1}{k} \mathbf{w}_{t_j}$. Next, a vertex $v_i \in V$ representing the event $e_i$ receives the vertex features $\mathbf{g}_{v_i} = \mathbf{g}_{e_i}$. These features are used in the graph regularization process.

The GNEE graph regularization framework has two assumptions. First, neighboring vertices must have similar feature vectors. Second, event feature vectors must remain unchanged during regularization. Equation \ref{eq:reg-gfhf} defines the objective function to be minimized for graph regularization. The first term is related to the first assumption, where two neighboring vertices with weight $w_{v_{i}, v_{j}}$ must have a low similarity difference between their feature vectors. The second term is related to the second assumption, where the event vertices must preserve their feature vectors. The term $ \lim_{\mu\to \infty}\mu$ guarantees that a small difference $(\mathbf{f}_{v_{i}}-\mathbf{g}_{v_{i}})^2$ greatly penalizes the objective function $Q(\mathbf{F})$.

\begin{equation}
Q(\mathbf{F})=\frac{1}{2}\sum_{v_{i}, v_{j}\in V} w_{v_{i}, v_{j}} (\mathbf{f}_{v_{i}}-\mathbf{f}_{v_{j}})^2 + \lim_{\mu\to \infty}\mu \sum_{v_{i}\in V_L}(\mathbf{f}_{v_{i}}-\mathbf{g}_{v_{i}})^2
\label{eq:reg-gfhf}
\end{equation}

Equation \ref{eq:reg-gfhf} is a particular case of graph regularization, which has theoretical proofs of convergence \cite{ref:Belkin2006,ref:Zhu2003}. It can be solved via minimization with quadratic programming or through iterative methods based on label propagation.

After the graph regularization step, all graph vertices will have features in the same event feature space $\mathbf{F}$, i.e., the vertex regularized features. Now, the next step of the GNEE is the semi-supervised graph embedding learning with attention mechanisms. The input of this step is a set of regularized vertex features $\mathbf{F} \in \mathbb{R}^{|V| \times m}$ (from the previous step), where $m$ is the dimension of the event textual features and $|V|$ is the total of vertices. GNEE aims to learn a new set of high-level features $\mathbf{Z} \in  \mathbb{R}^{|V| \times d}$, where $d$ is the dimension of the new learned space (graph embedding space). An innovation of GNEE in relation to the existing event analysis methods is to explore a shared self-attention mechanism  $att : \mathbb{R}^\mathbf{Z} \times \mathbb{R}^\mathbf{Z} \rightarrow \mathbb{R}$ proposed by \cite{velickovic2018graph}, defined in Equation \ref{att1}, where $\mathbf{A} \in \mathbb{R}^{d \times m}$ is a weight matrix, and $\mathbf{z}_i$ and $\mathbf{z}_j$ are feature vectors of the vertices $v_i$ and $v_j$, respectively.

\begin{equation} \label{att1}
    a_{v_i,v_j} = att(\mathbf{A}\mathbf{z}_{v_i}, \mathbf{A}\mathbf{z}_{v_j})
\end{equation}

An important step of the graph attention networks is to consider relationships between events and components into the attention mechanism. In this case, $a_{v_i,v_j}$ is only computed for the neighboring nodes $\mathcal{N}_{v_i}$ of the vertex $v_i$, followed by a normalization via softmax function, according to Equation \ref{attsoftmax}. In this equation, $\alpha_{v_i,v_j}$ indicates the normalized importance of the $v_j$ features to vertex $v_i$ considering the $k$ neighboring vertices $\mathcal{N}_{v_i}$.

\begin{equation} \label{attsoftmax}
    \alpha_{v_i,v_j}=\operatorname{softmax}\left(a_{v_i,v_j}\right)=\frac{\exp \left(a_{v_i,v_j}\right)}{\sum_{v_k \in \mathcal{N}_{v_i}} \exp \left(a_{v_i,v_k}\right)}
\end{equation}.

The attention coefficients $\alpha_{v_i,v_j} \forall v_j \in \mathcal{N}_{v_i}$ are used to learn the feature vector $\mathbf{z}_{v_i}$ through a linear combination from all neighboring vertex features, as defined in Equation \ref{attsingle}, where $\sigma$ represents some non-linearity function. This process is applied to all vertices, thus obtaining the graph embedding space $\mathbf{Z}$.

\begin{equation} \label{attsingle}
    \mathbf{z}_{v_i} = \sigma\left(\sum_{v_j \in \mathcal{N}_{v_i}} \alpha_{v_i,v_j} \mathbf{A} \mathbf{z}_{v_j}\right)
\end{equation}

Equation \ref{attsingle} represents the embedding calculation considering a single attention mechanism for the entire event graph. However, previous studies show that multiple attention mechanisms can learn more appropriate representations \cite{velickovic2018graph}, especially in graphs with complex structures. Thus, since an event graph is composed of different objects and relationships, our GNEE explores multiple and independent attention mechanisms. We argue that a minimum of $C$ attention mechanisms are sufficient to learn high-level event features, where $C$ represents the total of event components. The key idea is that each attention mechanism (hopefully) can learn the importance of each component vertices and their relationship to event vertices.

\begin{equation} \label{multiatt}
    \mathbf{z}_{v_i}= \bigparallel_{c=1}^{C} \sigma\left(\sum_{v_j \in \mathcal{N}_{v_i}} \alpha_{v_i,v_j}^{(c)} \mathbf{A}^{(c)} \mathbf{z}_{v_j}\right) 
\end{equation}

Equation \ref{multiatt} defines the GNEE multi-head attention mechanism. The $\bigparallel$ operator indicates the concatenation of the features obtained by each attention mechanism, where $\alpha_{v_i,v_j}^{(c)}$ is the attention coefficient of the $c$-th layer and the respective weight matrix $\mathbf{A}^{(c)}$.

\begin{figure}[htpb]
    \centering
  \includegraphics[width=0.7\linewidth]{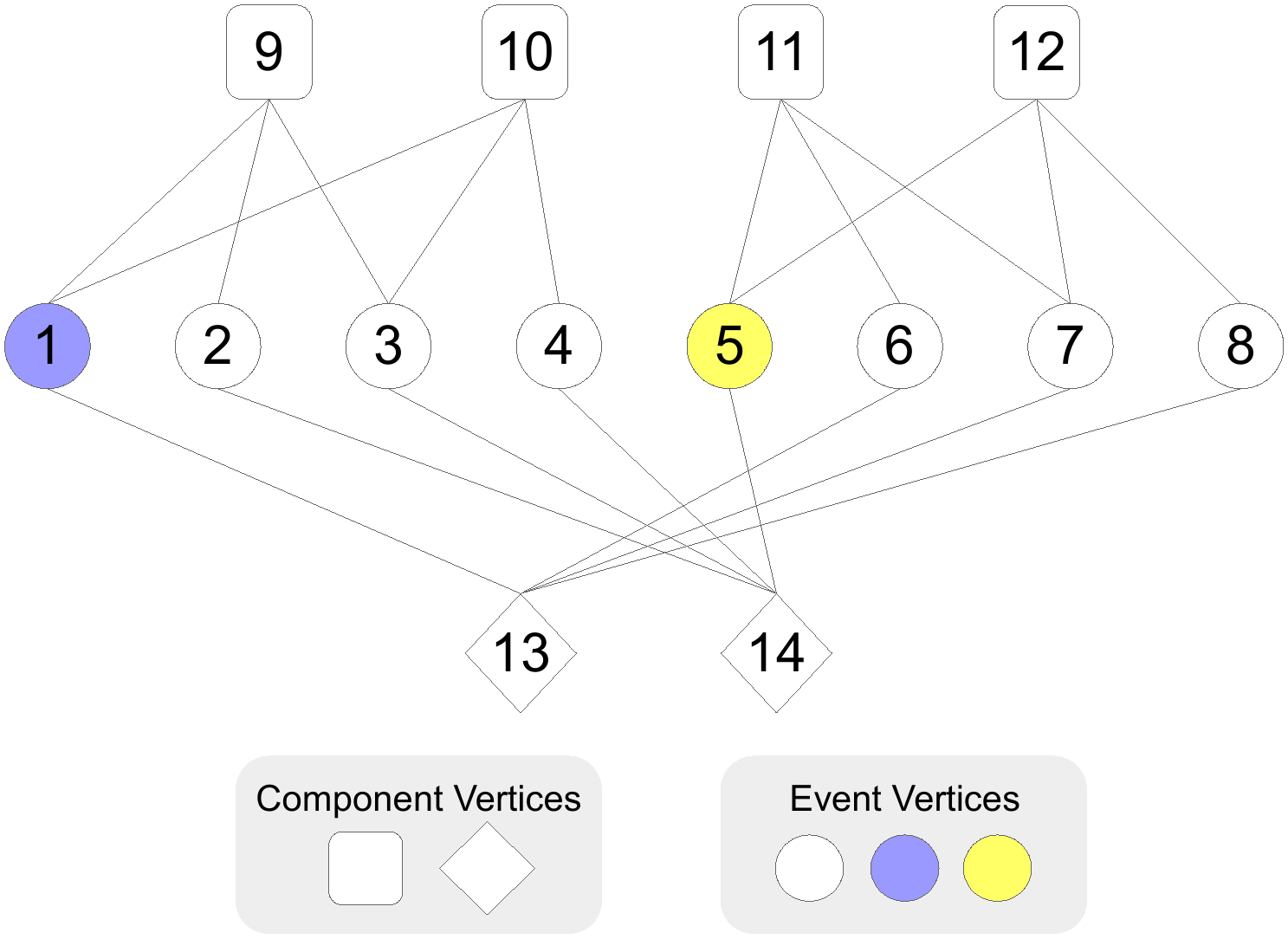}
  \caption{Example of an event graph containing 8 events (two labeled events) and 6 components of two types.}
  \label{fig:event_graph_gat_example}
\end{figure}

\begin{figure}[htpb]
    \centering
  \includegraphics[width=\linewidth]{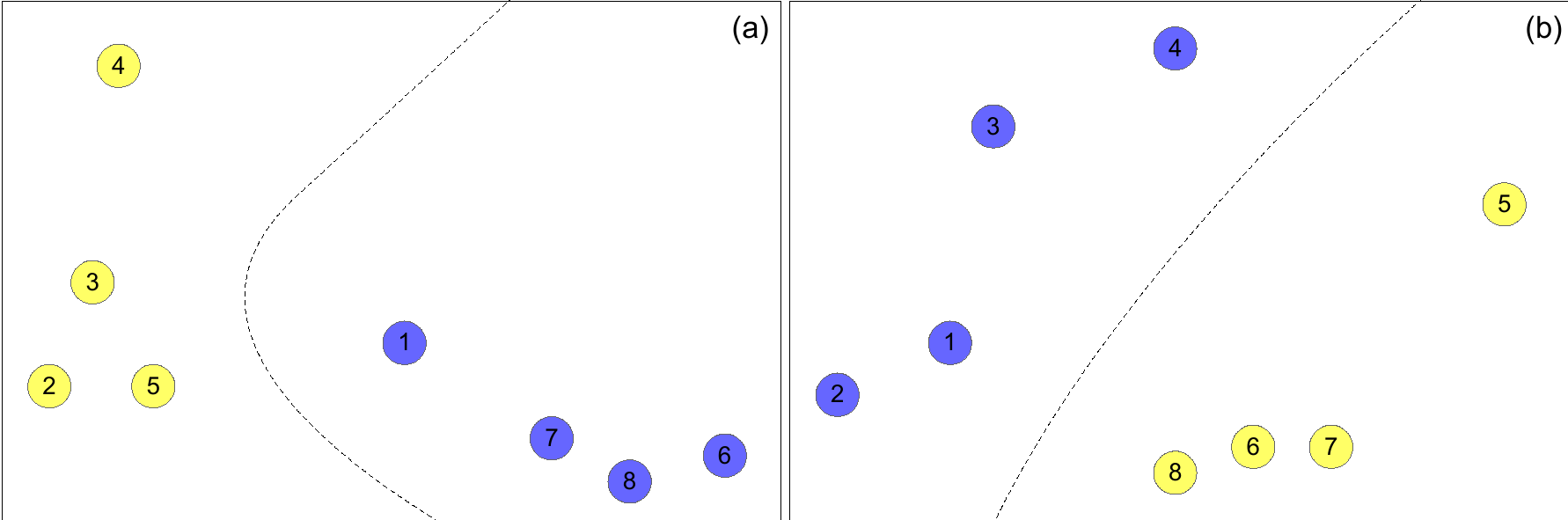}
  \caption{Latent spaces obtained by two attention matrices. Dashed lines indicate decision boundaries.}
  \label{fig:example_att_hin}
\end{figure}

To exemplify the GNEE multi-head attention mechanism, Figure \ref{fig:event_graph_gat_example} presents an event graph containing $8$ event vertices and $6$ component vertices of $2$ different types. For example, consider that there are $4$ vertices of organizations and $2$ geographic vertices. The colored vertices represent labeled events. Note that if we only consider organization vertices, then events can be classified as $\{1,2,3,4\}$ and $\{5,6,7,8\}$. On the other hand, if we consider only geographic vertices, events can be classified as $\{1,6,7,8\}$ and $\{2,3,4,5\}$. We run GNEE with two attention matrices and two hidden layers, where each attention matrix learns two latent features. Figures \ref{fig:example_att_hin}a and \ref{fig:example_att_hin}b show the embedding space learned by each attention matrix, which was able to properly identify how each event component acts in the classification problem (see decision boundaries). The GNEE source code to replicate this example, as well as to extract features from each attention mechanism, are available at \url{https://github.com/joaopedromattos/GNEE}.

Finally, we highlight some GNEE capabilities concerning exploratory event analysis tasks. In GNEE, both confidence classification vectors and feature vectors (embeddings) are available for all graphs' vertices. Thus, we can explore how important a specific component is for a given class, as well as calculate the similarity between pairs of events, pairs of components, and between events and components. In the next section, we discuss GNEE performance to learn event graph embeddings and present an experimental comparison involving other state-of-the-art graph embeddings methods.

\section{Experimental Evaluation}

\subsection{Experimental Setup and Baselines}

We carried out an experimental evaluation involving $5$ real-words event datasets \cite{Hamborg2019b}.  Table \ref{dataset_overview} shows an overview of each dataset, including the total of vertices, edges, the average degree of vertices, and the number of classes.

\begin{table}[htbp]
\centering
\caption{Overview of the event graphs used in the experimental evaluation.}
\begin{tabular}{l|c|c|c|c}
\hline
\hline
Event Graph & $|V|$ & $|E|$ & Avg. Degree & \#Classes \\ \hline
\hline
GoogleNews & 227 & 270 & 2.38 & 7 \\ \hline
BBC & 392 & 453 & 2.31 & 5 \\ \hline
GoldStd & 579 & 803 & 2.77 & 13 \\ \hline
CLNews & 2191 & 3208 & 2.92 & 69 \\ \hline
40ER & 249 & 344 & 2.76 & 3 \\ \hline
\hline
\end{tabular}
\label{dataset_overview}
\end{table}

Our GNEE neural networks was configured with $8$ attention matrices and $8$ hidden layers. Each attention mechanism derives $8$-dimensional embeddings. After the concatenation step, the graph-embedding space will consist of $64$-dimensional feature vectors. 
We used the DistilBERT Multilingual model from SentenceTransformers tool\footnote{\url{https://github.com/UKPLab/sentence-transformers}} to generate features from event texts. These features were used in the event feature regularization step (Equation \ref{eq:reg-gfhf}). We compared our GNEE with $5$ graph embedding methods: DeepWalk, Node2Vec, Struct2Vec, GCN, and GAT.

\begin{table*}[htpb]
\centering
\caption{F1 classification performance values obtained from the embeddings of each method}
\begin{tabular}{l|c|c|c|c|c}
\hline
\hline
 & 40ER & BBC & GoldStd & GoogleNews & CLNews \\ \hline
 \hline
DeepWalk & 0.629 ± 0.09 & 0.404 ± 0.05 & 0.508 ± 0.05 & 0.622 ± 0.09 & 0.507 ± 0.03 \\ \hline
GAT & 0.594 ± 0.06 & 0.377 ± 0.07 & 0.480 ± 0.04 & 0.506 ± 0.10 & 0.475 ± 0.02 \\ \hline
GCN & 0.638 ± 0.10 & 0.458 ± 0.09 & 0.548 ± 0.04 & 0.617 ± 0.09 & \textbf{0.519 ± 0.03} \\ \hline
Node2Vec & 0.630 ± 0.11 & 0.428 ± 0.06 & 0.546 ± 0.05 & 0.584 ± 0.08 & 0.508 ± 0.02 \\ \hline
Struct2Vec & 0.415 ± 0.07 & 0.186 ± 0.05 & 0.089 ± 0.02 & 0.311 ± 0.05 & 0.066 ± 0.00 \\ \hline
GNEE (ours) & \textbf{0.744 ± 0.12} & \textbf{0.634 ± 0.07} & \textbf{0.676 ± 0.03} & \textbf{0.958 ± 0.07} & 0.274 ± 0.02 \\ \hline
\hline
\end{tabular}
\label{res1}
\end{table*}

\begin{table*}[htpb]
\centering
\caption{ACC classification performance values obtained from the embeddings of each method}
\begin{tabular}{l|c|c|c|c|c}
\hline
\hline
 & \multicolumn{1}{c|}{40ER} & \multicolumn{1}{c|}{BBC} & \multicolumn{1}{c|}{GoldStd} & \multicolumn{1}{c|}{GoogleNews} & \multicolumn{1}{c}{CLNews} \\ \hline
 \hline
DeepWalk & 0.751 ± 0.07 & 0.415 ± 0.05 & 0.638 ± 0.05 & 0.637 ± 0.09 & 0.607 ± 0.03 \\ \hline
GAT & 0.775 ± 0.06 & 0.393 ± 0.05 & 0.609 ± 0.04 & 0.548  ± 0.10 & 0.568 ± 0.02 \\ \hline
GCN & 0.756 ± 0.10 & 0.473 ± 0.06 & 0.655 ± 0.05 & 0.628 ± 0.12 & 0.601 ± 0.02 \\ \hline
Node2Vec & 0.751 ± 0.09 & 0.439 ± 0.06 & 0.669 ± 0.05 & 0.620 ± 0.08 & \textbf{0.609 ± 0.03} \\ \hline
Struct2Vec & 0.530 ± 0.08 & 0.203 ± 0.04 & 0.117 ± 0.03 & 0.354 ± 0.08 & 0.081 ± 0.00 \\ \hline
GNEE (ours) & \textbf{0.778 ± 0.10} & \textbf{0.636 ± 0.06} & \textbf{0.795 ± 0.02} & \textbf{0.968 ± 0.05} & 0.420 ± 0.02 \\ \hline
\end{tabular}
\label{res2}
\end{table*}

In the GNEE experimental evaluation, we randomly selected $20$\% of event vertices as labeled vertices, thereby simulating a semi-supervised learning scenario. After the graph embedding step, the rest of the unlabeled events are classified considering the embeddings as input for a final layer with a logistic sigmoid activation. In order to evaluate the experimental results, we used the F1-Macro (F1) and Accuracy (ACC) measures \cite{Manning2008}. The same semi-supervised scenario is used to evaluate the GCN and GAT methods. However, the event feature regularization step was not used for these two methods. DeepWalk, Node2Vec, and Struct2Vec methods learn embeddings in an unsupervised way. Then the labeled events are used only in the classification step. In this case, we used the Support Vector Machine with a linear kernel.

\subsection{Results and Discussion}

We analyze and discuss the experimental results considering two aspects. First, we present the F1 and ACC values achieved by GNEE in comparison with the existing methods. Second, we perform a visual comparison of the embeddings obtained by each method using a two-dimensional projection of the embeddings, thereby allowing to qualitatively compare the graph representation learning.

Tables \ref{res1} and \ref{res2} show the classification performance considering F1 and ACC measures, respectively. For both measures, GNEE achieved the best performance in four out of five datasets.
In the datasets in which GNEE achieved better performance, GNEE showed a minimum improvement of $18$\% in the F1 measure and $3$\% in the ACC measure.

\begin{figure}[htbp]
    \centering
  \includegraphics[width=\linewidth]{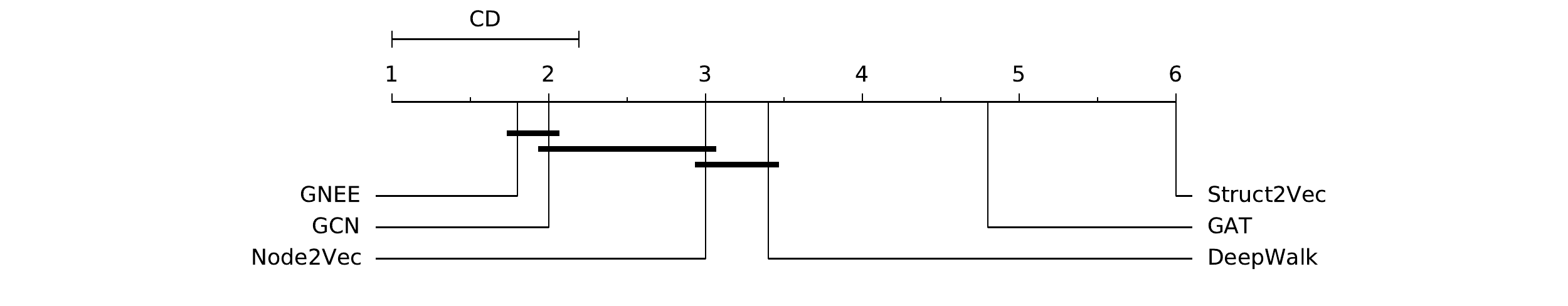}
  \caption{Critical difference diagram for the F1 measure.}
  \label{fig:eval1}
\end{figure}

\begin{figure*}[htpb] 
    \centering
    \begin{subfigure}[t]{0.31\textwidth}
        \centering
        \includegraphics[width=\linewidth]{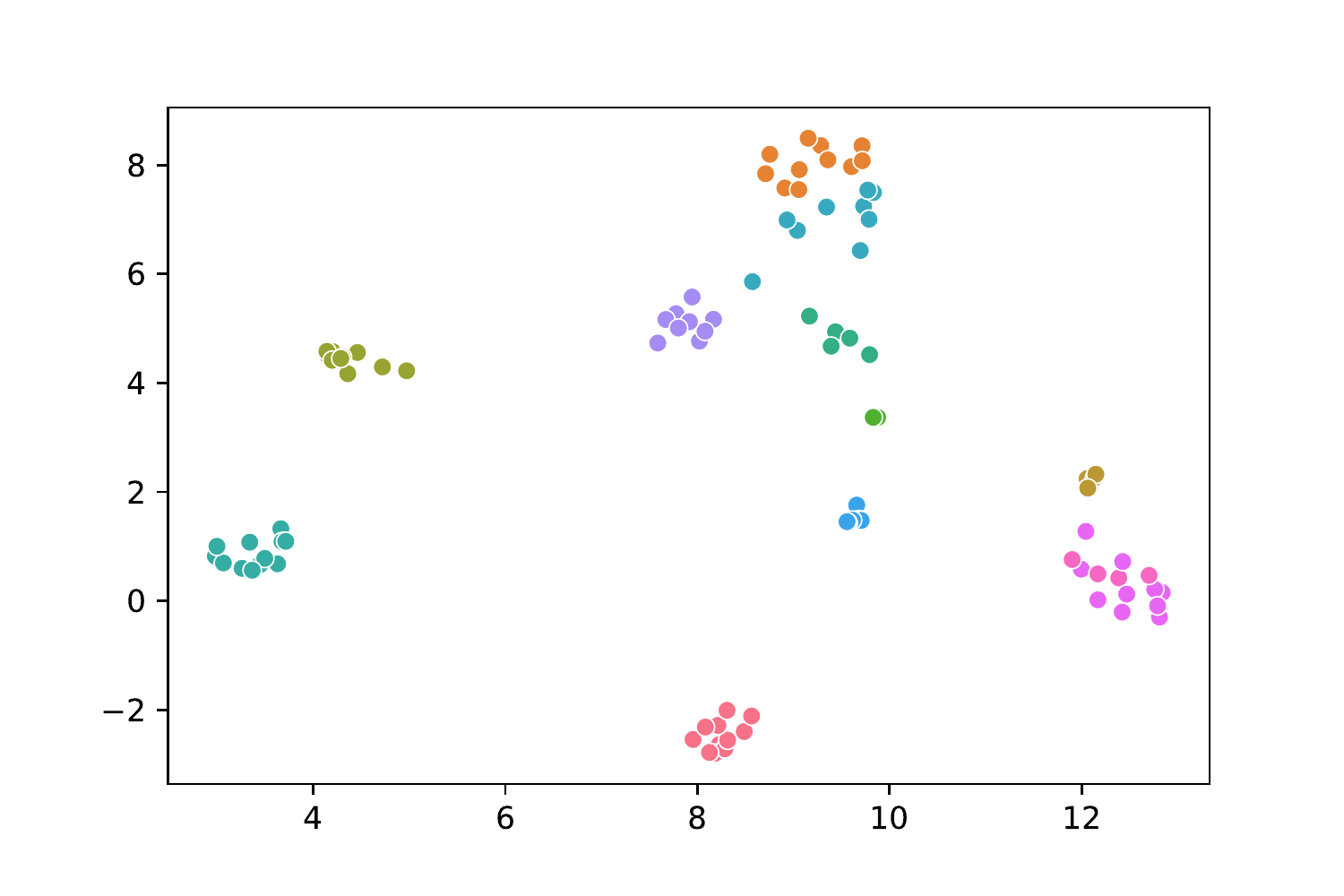}
        \caption{GNEE}
    \end{subfigure}%
    ~ 
    \begin{subfigure}[t]{0.31\textwidth}
        \centering
        \includegraphics[width=\linewidth]{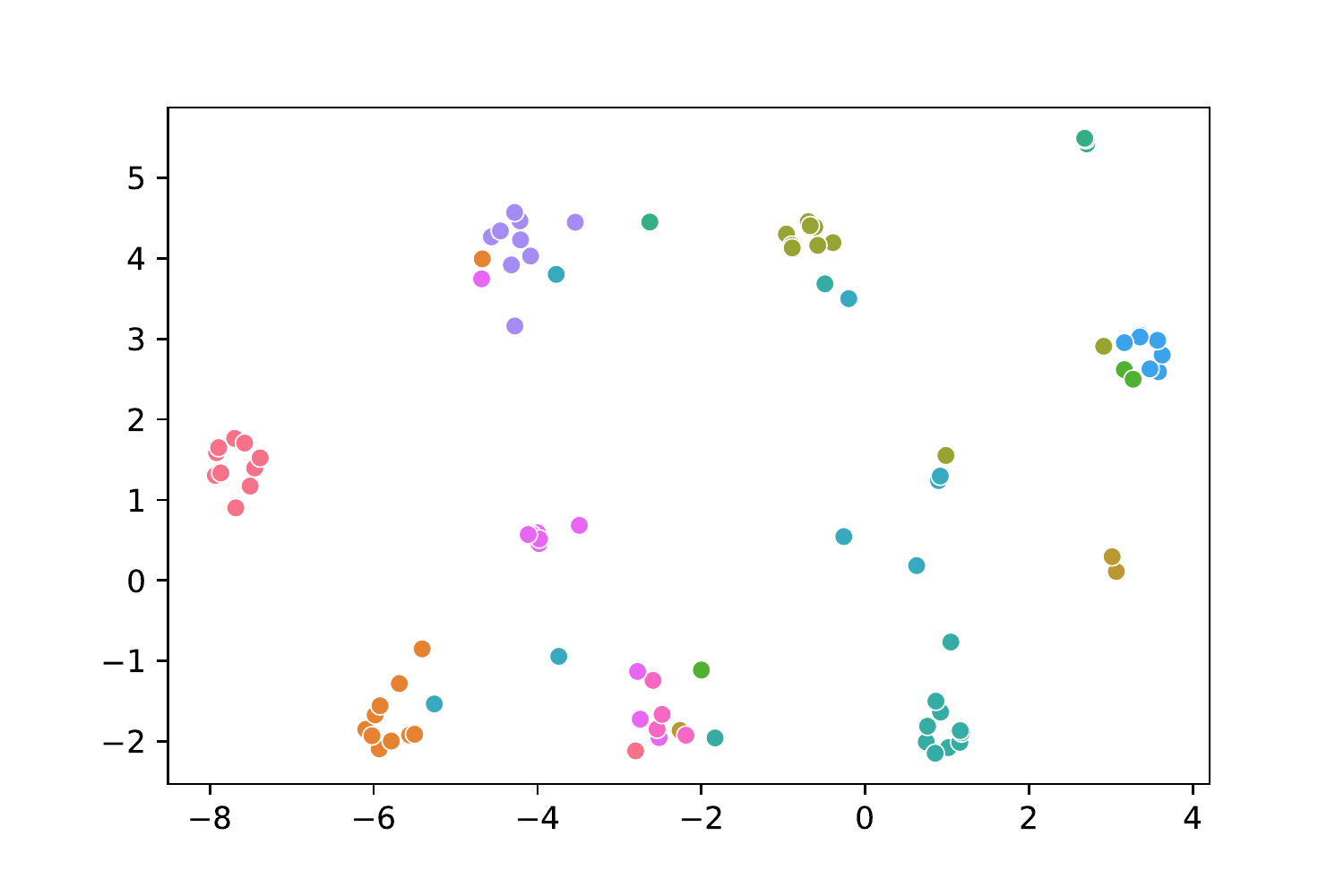}
        \caption{GCN}
    \end{subfigure}
    ~ 
    \begin{subfigure}[t]{0.31\textwidth}
        \centering
        \includegraphics[width=\linewidth]{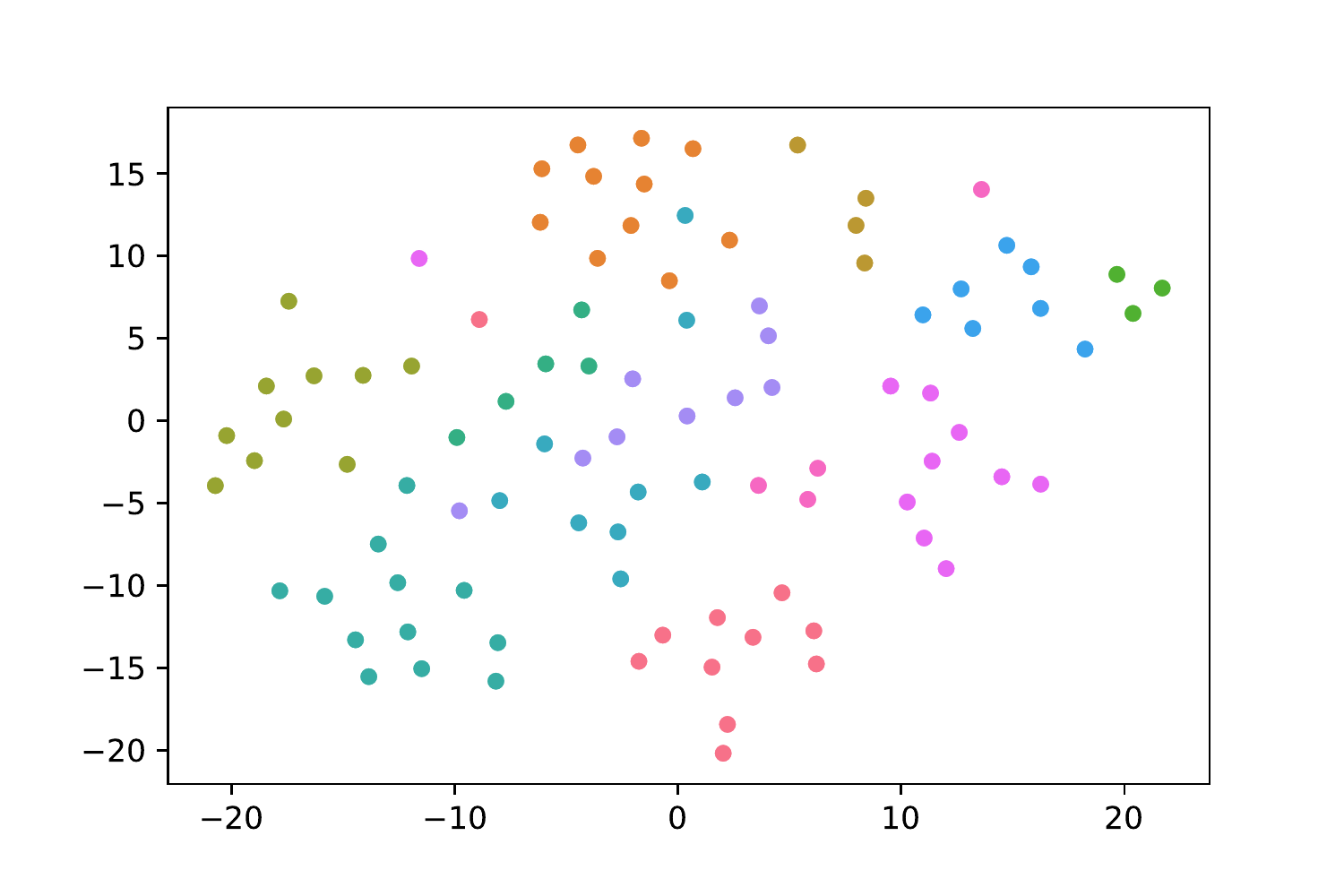}
        \caption{Node2Vec}
    \end{subfigure}
    ~ 
    \begin{subfigure}[t]{0.31\textwidth}
        \centering
        \includegraphics[width=\linewidth]{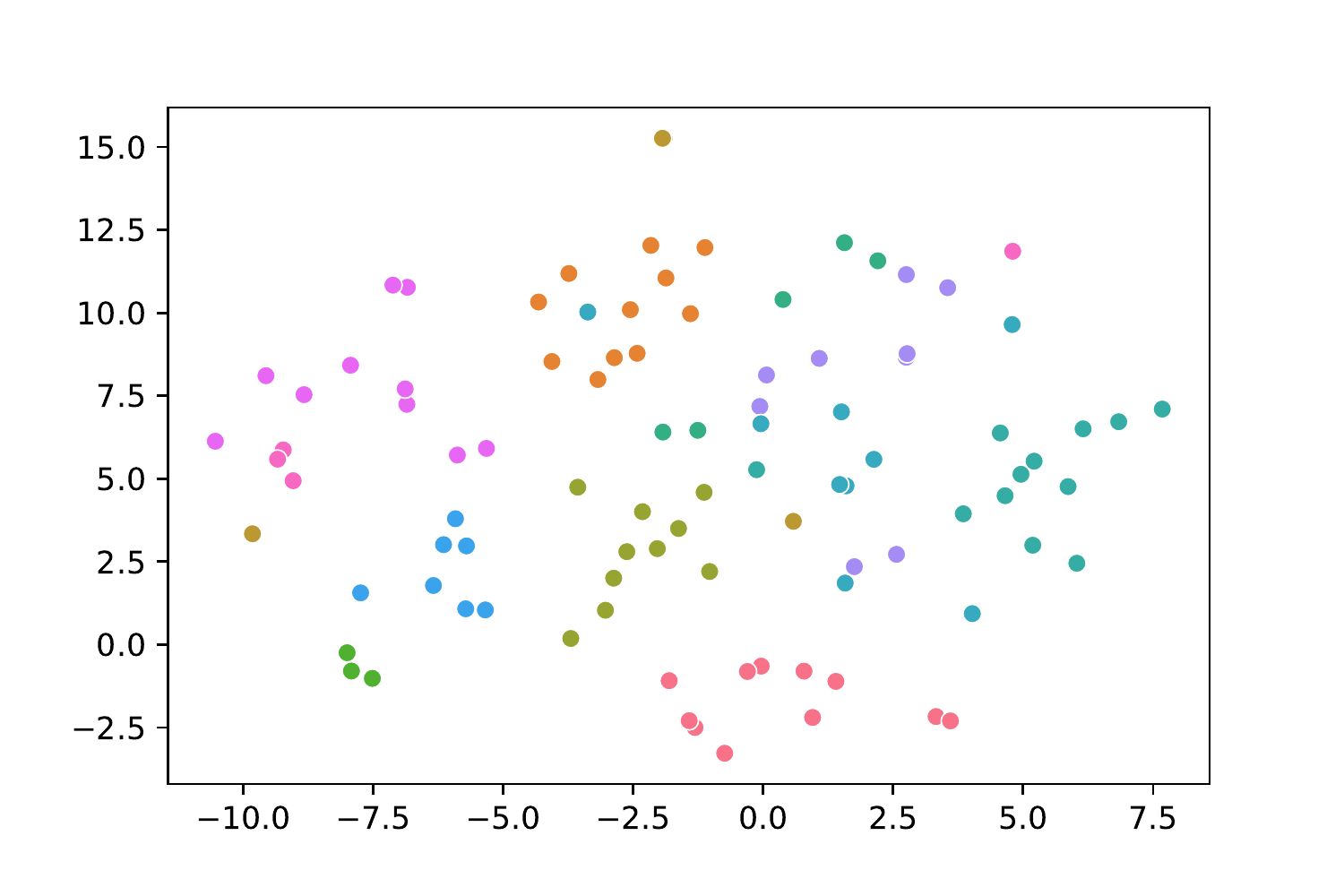}
        \caption{DeepWalk}
    \end{subfigure}
    ~ 
    \begin{subfigure}[t]{0.31\textwidth}
        \centering
        \includegraphics[width=\linewidth]{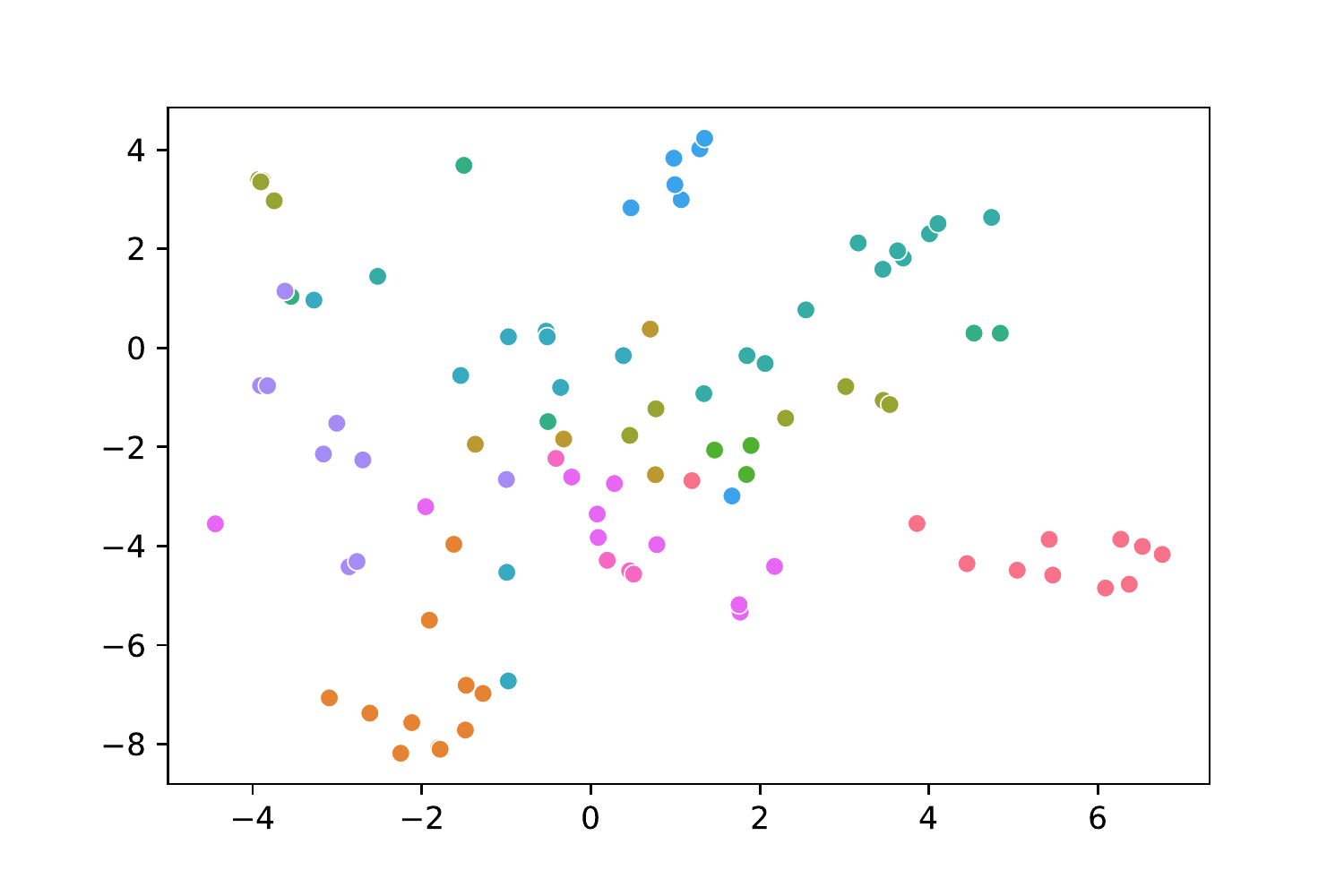}
        \caption{GAT}
    \end{subfigure}
    ~ 
    \begin{subfigure}[t]{0.31\textwidth}
        \centering
        \includegraphics[width=\linewidth]{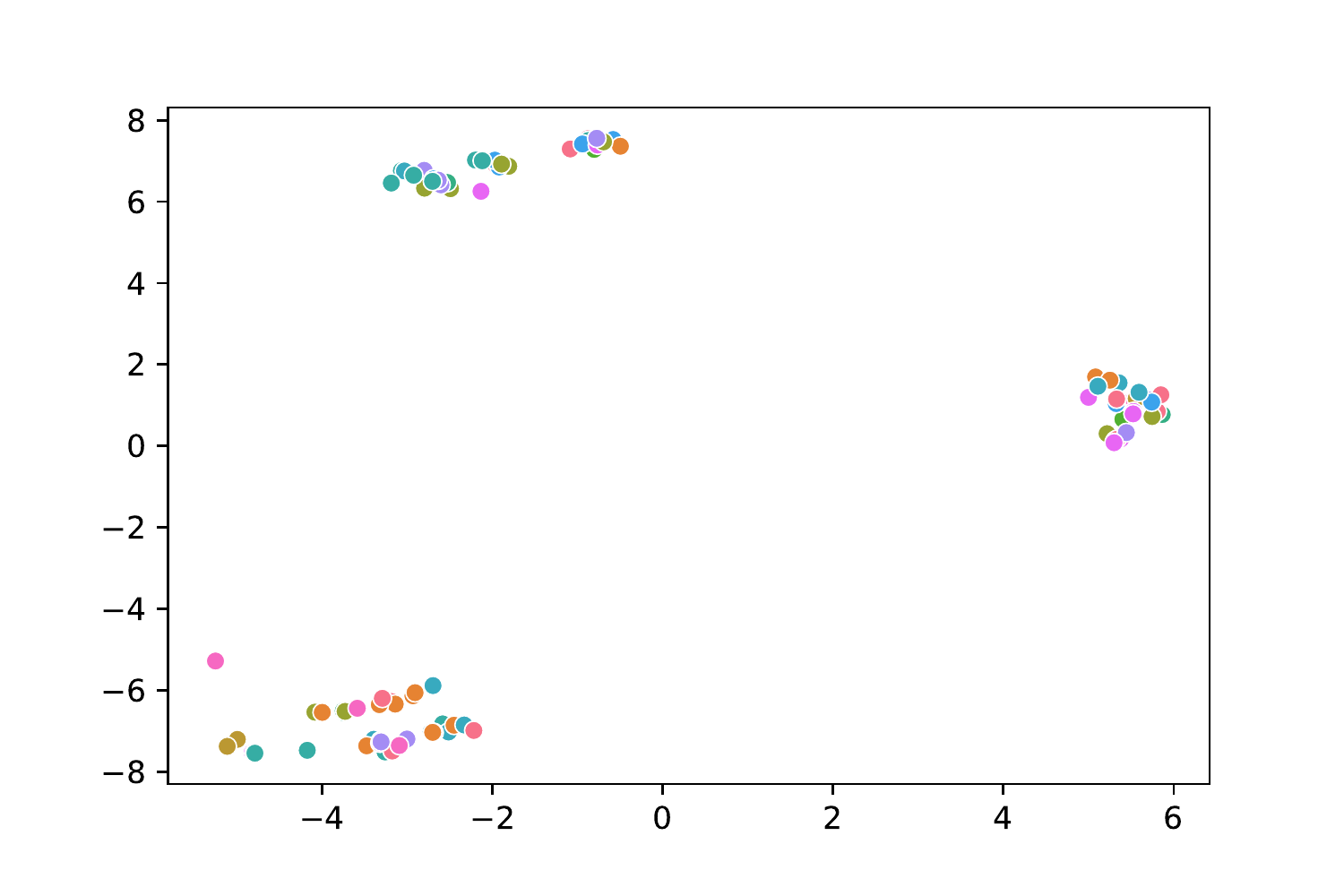}
        \caption{Struc2Vec}
    \end{subfigure}
    \caption{Two-dimensional projection (t-SNE) from the embeddings of each method in the GoldStd event graph.}
    \label{embstsne}
\end{figure*}

An important analysis is the comparison of GNEE and traditional GAT, in which we can compare the performance improvement obtained by the event feature regularization proposed in GNEE for event graphs. In the four datasets in which the GNEE obtained the best results, the event feature regularization led to improvements between 25\% to 70\% of the classification performance in both F1 and ACC measures. This result highlights the advantages of GNEE in incorporating textual event features for all vertices during graph embedding.

Figure \ref{fig:eval1} shows the critical difference diagram for F1 measure, computed by Friedman's test with Nemenyi's post-test with 95\% of confidence level. The methods are ordered considering the average ranking from multiple executions. We connect two methods with a line if there is no statistically significant difference between them. Although GNEE obtained the first position in the ranking, there is no evidence of statistically superior performance in relation to the GCN and Node2Vec methods. However, GNEE statistically outperforms the GAT method, thereby favoring the event feature regularization. Node2Vec and DeepWalk present similar performance considering several event graphs. However, Node2Vec allows the adjustment of the walking bias and, consequently, selecting the best models according to the event graph.

Struc2Vec was unable to learn suitable embeddings for event graphs. In fact, Struc2Vec was hampered by using the highest degree component vertices to learn embedding space due to its representation learning bias based on the graph's structural identity. We argue that Struc2Vec can be improved for event graphs if we restrict the search for structural identity using only events as the ``target vertices'' of the structure.

A second aspect of the experimental discussion is to visually analyze the embedding space learned by each method. We selected the GoldStd graph for this analysis. We use the t-SNE algorithm to project embeddings from 64-dimensions to a two-dimensional space, as shown in Figure \ref{embstsne}. In addition, we color each event according to its label. Note that the separation of events according to their classes is visually perceptible in GNEE, as well as in GCN, indicating suitable representation of the learned embedding space. Node2Vec and DeepWalk present an intersection of events of different categories, which is expected since they are unsupervised methods. In this dataset, GAT was inefficient in learning embeddings in the absence of vertice features.

\section{Conclusion}

We present, discuss and evaluate state-of-the-art methods for representation learning from event graphs. Moreover, we present important requirements for event representation learning: (i) allowing semi-supervised graph embedding to consider some labeled data; (ii) automatically learning the importance of event components; and (iii) dealing with the absence of features for some vertices.

We propose the GNEE (GAT Neural Event Embeddings) method that meets the requirements presented above and obtains competitive performance compared to the existing methods. GNEE incorporates state-of-the-art techniques in graph learning to perform an event feature regularization and mitigate the challenge of learning embeddings from graphs with events and their components. Then, attention mechanisms are used to determine the importance of a vertex and its neighbors, which helps determine the importance of events and components. The GNEE source code is available in \url{https://github.com/joaopedromattos/GNEE}, as well as the datasets and source code to reproduce the experiments.

Directions for future work involve investigating how the attention mechanisms act on event graphs using an explainable AI methodology. The general idea is to incorporate the recent advances in interpretable models for event representation learning from graphs.

\bibliographystyle{IEEEtran}

\bibliography{bare_conf}

\end{document}